\documentclass{article}
\usepackage[utf8]{inputenc}
\usepackage{amsmath, amssymb}
\usepackage{graphicx}
\usepackage{geometry}
\usepackage{hyperref}
\usepackage{authblk}
\usepackage{graphicx}
\usepackage{caption}
\usepackage{subcaption}
\usepackage[table]{xcolor}

\title{Radius-Guided Post-Clustering for Shape-Aware, Scalable Refinement of k-Means Results}
\author{
	Stefan Kober \\
	Independent Researcher \\
	\texttt{stefan.kober@ieee.org} \\
	\url{https://orcid.org/0000-0003-1319-8282}
}

\date{\today}

\begin{document}

\maketitle

\begin{abstract}
Traditional k-means clustering underperforms on non-convex shapes and requires the number of clusters $k$ to be specified in advance. We propose a simple geometric enhancement: after standard k-means, each cluster center is assigned a radius (the distance to its farthest assigned point), and clusters whose radii overlap are merged. This post-processing step loosens the requirement for exact $k$: as long as $k$ is overestimated (but not excessively), the method can often reconstruct non-convex shapes through meaningful merges.
	
We also show that this approach supports recursive partitioning: clustering can be performed independently on tiled regions of the feature space, then globally merged, making the method scalable and suitable for distributed systems. 

Implemented as a lightweight post-processing step atop scikit-learn’s k-means, the algorithm performs well on benchmark datasets, achieving high accuracy with minimal additional computation.
\end{abstract}

\section{Introduction}

K-means clustering remains one of the most widely used unsupervised learning methods due to its simplicity, speed, and intuitive geometric structure. However, its core limitations are well known: it requires the number of clusters $k$ to be specified in advance, and it performs poorly when clusters have irregular shapes or are not linearly separable~\cite{steinley2006kmeans, JAIN2010651, Validity, KinKmeans}.

Many enhancements to k-means have been proposed, ranging from improved initialization strategies such as k-means++~\cite{Arthur} to variations in objective functions and distance metrics. Other clustering algorithms—such as DBSCAN~\cite{Ester}, hierarchical clustering~\cite{muellner2011}, and Gaussian mixture models~\cite{bishop2006pattern}—aim to overcome k-means' limitations with respect to shape and distributional assumptions, but often at the cost of increased computational complexity or reduced interpretability.

This paper introduces a simple geometric post-processing enhancement to standard k-means. Each cluster center is assigned a \emph{radius}, defined as the distance to its farthest assigned point. After initial clustering, two clusters are merged if the sum of their radii exceeds the Euclidean distance between their centers. Intuitively, this means their radius-based boundaries would overlap if drawn around their respective centroids.

\begin{figure}[htbp]
	\centering
	
	\begin{subfigure}{0.3\textwidth}
		\includegraphics[width=\linewidth]{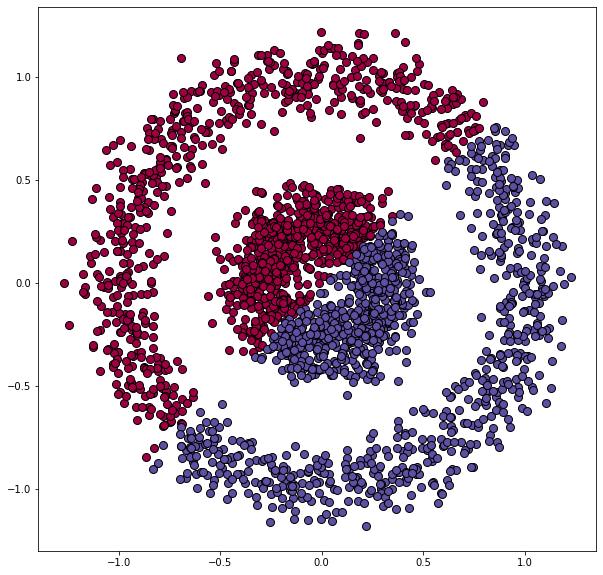}
		\caption{k-means with $k=2$}
	\end{subfigure}
	\hfill
	\begin{subfigure}{0.3\textwidth}
		\includegraphics[width=\linewidth]{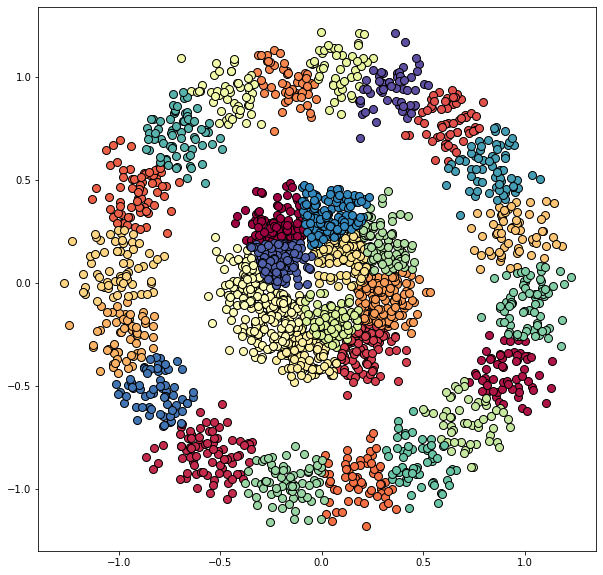}
		\caption{k-means with $k=20$}
	\end{subfigure}
	\hfill
	\begin{subfigure}{0.3\textwidth}
		\includegraphics[width=\linewidth]{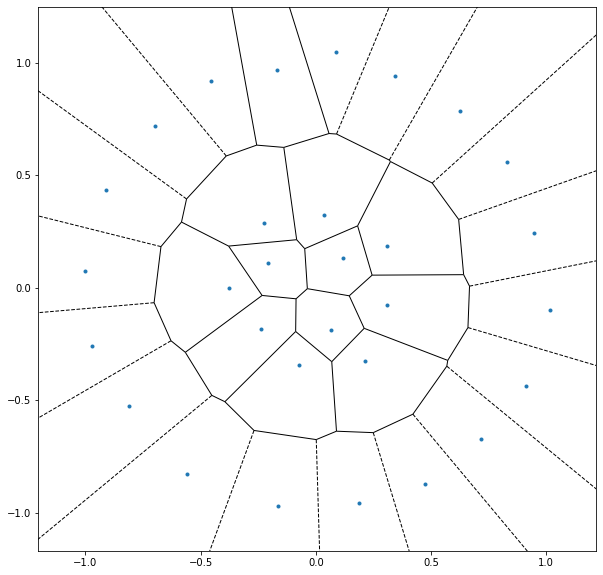}
		\caption{Induced Voronoi cells ($k=20$)}
	\end{subfigure}
	
	\vspace{0.5em}
	
	\begin{subfigure}{0.3\textwidth}
		\includegraphics[width=\linewidth]{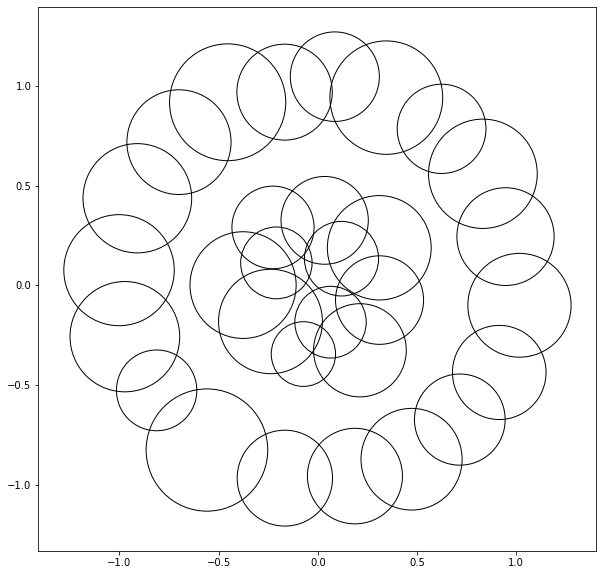}
		\caption{Circles around cluster centers ($k=20$)}
	\end{subfigure}
	\hfill
	\begin{subfigure}{0.3\textwidth}
		\includegraphics[width=\linewidth]{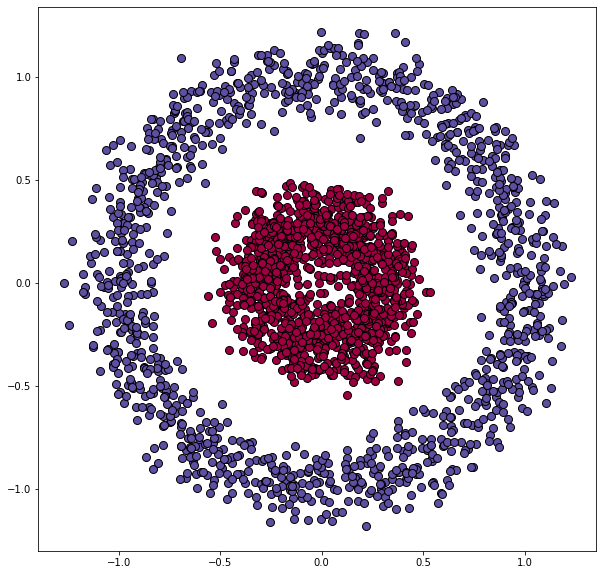}
		\caption{Post-merge result based on (d)}
	\end{subfigure}
	
	\caption{
		Visual explanation of radius-based post-processing.  
		\textbf{(a)} Naive k-means with $k=2$ fails to recover the circular structure.  
		\textbf{(b)} With $k=20$, clusters oversegment the shape.  
		\textbf{(c)} Voronoi cells offer no guidance for merging.  
		\textbf{(d)} Circles around cluster center through farthest point reveal meaningful overlap.  
		\textbf{(e)} The correct clustering is recovered by radius-based merging of the clusters in (b).
	}
	\label{fig:circle-explainer}
\end{figure}

The underlying assumption of this heuristic is that merging based on radius overlap preserves the intrinsic structure of the data. In standard k-means, each cluster implicitly defines a Voronoi cell in the feature space, but these cells offer no principled criterion for merging. By assigning each cluster a radius, we introduce a circular structure around each center—enabling a geometric merge rule based on potential overlap. This allows the algorithm to recover from an overestimated $k$ and to progressively unite adjacent regions that likely belong to the same underlying cluster. Figure~\ref{fig:circle-explainer} illustrates this progression on a classic two-circle dataset, showing how standard k-means fails and how radius-guided merging restores the ground truth labels.

The method has been implemented as a lightweight extension atop scikit-learn’s standard k-means algorithm. It relies solely on standard information already computed during the k-means process—namely, cluster centers and point assignments—making it straightforward to adapt to other implementations as well.

The method performs well on synthetic datasets, including classic shape-based examples like two moons and concentric circles, as well as the more challenging benchmarks from the Fundamental Clustering Problem Suite (FCPS), introduced in~\cite{ultsch} and available from~\cite{FCPS_Repo}. Across a wide range of initial $k$ values, the method consistently achieved median clustering success rates of 98--100\%, with low interquartile variability. These results suggest that the radius-based merge rule is highly stable and effective in noise-free conditions, even when the initial k-means partitioning significantly oversegments the data.

In moderately noisy settings, clusters with radius zero—typically corresponding to isolated outliers—can be excluded prior to merging. While this heuristic offers a degree of robustness, it is not explored further in this paper. The method is not designed for high-noise environments, but this filtering step illustrates how noise handling could be extended without complicating the core algorithm.

The approach also supports recursive scalability. When the feature space is divided into rectangular tiles, clustering can be performed locally within each tile. Because the computed radii extend across tile boundaries, the post-processing merge step can naturally reconnect fragments of clusters that were split by partitioning. This makes the method well suited for large-scale or distributed clustering tasks.

This paper presents the algorithm, its geometric motivation, implementation details, and experimental results. The method offers a conceptually simple yet effective enhancement to k-means—improving shape recovery and scalability while preserving interpretability and speed.

\section{Related Work}

Several clustering methods have been proposed to address the limitations of k-means—particularly its limited performance on non-convex clusters, and its reliance on a fixed number of clusters $k$.

Density-based approaches such as DBSCAN~\cite{dbscan} and HDBSCAN~\cite{hdbscan} are well known for their ability to identify clusters of arbitrary shape and automatically determine the number of clusters. However, these methods depend on density thresholds and reachability parameters that can be difficult to tune, especially in high dimensions or when data densities vary significantly. They are also sensitive to noise and can struggle with cluster separation in sparse or uneven regions, despite their ability to reject noise.

Hierarchical clustering methods~\cite{muellner2011} construct nested cluster trees, offering flexible granularity. However, they are often computationally intensive and lack a principled strategy for selecting the appropriate level of the hierarchy. Other approaches, such as Gaussian Mixture Models~\cite{bishop2006pattern}, provide probabilistic interpretations but rely on stronger distributional assumptions and perform poorly on non-elliptical clusters.

Several extensions to k-means, including k-means++~\cite{Arthur}, x-means~\cite{Xmeans} and KHM~\cite{alternatives}, improve initialization or offer limited adaptability in $k$, but they retain the core geometric constraints of the original algorithm. 

The method proposed in this paper stays close to the simplicity and efficiency of standard k-means while improving its adaptability to irregular shapes and over-segmented results. This over-segmentation, in turn, reduces the method's reliance on an exact choice of $k$. Unlike density-based or hierarchical models, it requires no density estimation, distance thresholds, or probabilistic assumptions. Instead, it introduces a simple geometric post-processing rule based on cluster radii—a lightweight enhancement that preserves the intuitive structure of k-means while expanding its flexibility and scalability.

\section{Algorithm Description}

\subsection{Standard k-means (Recap)}
The k-means algorithm partitions $n$ observations into $k$ clusters by minimizing the within-cluster sum of squares. It works by initializing $k$ centroids, assigning each point to its nearest centroid, and updating the centroids to the mean of their assigned points. This process repeats until convergence, typically measured by centroid stability or a fixed number of iterations.

However, k-means assumes that clusters are convex and roughly spherical. It also requires $k$ to be specified in advance. 

\subsection{Radius-Based Post-Clustering}
To address these limitations, we introduce a simple geometric enhancement. After running standard k-means (with an intentionally large $k$), we compute a \textit{radius} for each cluster $C_i$ as the maximum distance from its centroid $c_i$ to any of its assigned points:

\[
r_i = \max_{x \in C_i} \|x - c_i\|_2
\]

We then iteratively merge pairs of clusters $(C_i, C_j)$ whose centers are sufficiently close, according to the following merge condition:

\[
\|c_i - c_j\|_2 \leq r_i + r_j
\]

That is, two clusters are merged if their radius-based boundaries would overlap, assuming a spherical spread around each centroid.

\subsection{Post-Processing Implementation}
\label{sec:post-processing}

The merging step is implemented as a post-processing pass after k-means clustering. The initial value of $k$ can be chosen conservatively high; over-partitioned regions are corrected by merging overlapping clusters. However, if $k$ is chosen too high, the method may begin to interpret arbitrary local variation as meaningful boundaries. In the extreme case where $k = n$, the result is a micro-cluster for each data point, and merging cannot recover meaningful structure.

This sensitivity can be leveraged as a heuristic to estimate the number of natural clusters. If the data contain coherent clusters, their number should remain stable across a range of slightly overestimated $k$ values. A domain-aware strategy might evaluate post-processed results for $k \in \{2, 4, 6, 8, \dots\}$: initially, all points may fall into a single cluster; with larger $k$ more and more clusters are added; then a stable phase appears, during which the same number of clusters is returned, and their members overlap significantly; finally, as $k$ increases further, the number of resulting clusters rises and overlap decreases. This progression is akin to observing the data at different resolutions.

The merging process is implemented as a graph-based post-processing step. After k-means clustering, each cluster center is assigned a radius, and an adjacency matrix is constructed: two clusters are considered adjacent if their centers are closer than the sum of their radii. This defines a connectivity graph over the set of cluster centers. Clusters belonging to the same connected component in this graph are merged, and the new labels are propagated accordingly. 

This process avoids iterative pairwise merging and instead identifies all merged clusters in a single pass. It introduces minimal computational overhead and relies only on standard outputs of k-means: cluster centers and point assignments. The method can thus be added to most k-means pipelines with minimal modification.

A reference implementation of the full method, based on scikit-learn, is available in the open-source repository~\cite{radiusguidedkmeans2024}.

\section{Recursive Partitioning for Scalability}

One key advantage of the radius-based merging rule is that it enables clustering to be performed locally and then globally refined—making the method naturally scalable to large datasets.

The core insight is geometric: when a cluster is split across a boundary (e.g., by dividing the feature space into rectangular tiles), the radius of each partial cluster will usually extend beyond its tile. During post-processing, the radius-based merge rule can reconnect such fragments, effectively restoring clusters that were artificially separated by partitioning.

The enhanced algorithm for tiling and merging proceeds as follows:
\begin{enumerate}
	\item Divide the feature space into a regular grid of rectangular tiles.
	\item Independently run standard k-means on each tile. To ensure consistency across tiles with varying data densities, set $k$ as a proportion of each tile's number of data points.
	\item Apply the radius-based merging rule locally within each tile. Cluster centers with radii extending beyond their tile boundaries are recorded for global merging.
	\item In the global merging pass, merge cluster centers whose radii crossed tile boundaries using the same radius-based criterion. Computational efficiency can be improved by pre-excluding impossible pairings through spatial filtering strategies.
\end{enumerate}

This tile-aware strategy offers several advantages: it enables parallel or distributed clustering across subsets of large datasets. It gracefully handles datasets that exceed memory or compute limits on a single machine. And it preserves cluster continuity across tile boundaries without requiring global knowledge during initial clustering.

These properties make the method well suited for scalable clustering pipelines, including streaming data, tiled image domains, and large-scale geospatial or sensor data.

\section{Experimental Results}

To evaluate the proposed method, we tested it on a range of synthetic and structured datasets, namely the diverse benchmarks from the Fundamental Clustering Problem Suite (FCPS), introduced in~\cite{ultsch} and available from~\cite{FCPS_Repo}

The goal was to assess the algorithm’s ability to:
\begin{itemize}
	\item Recover irregular or non-convex cluster shapes
	\item Merge over-segmented clusters caused by large initial $k$
	\item Maintain high overlap with labeled ground truth
\end{itemize}

\begin{figure}[htbp]
	\centering
	
	\begin{subfigure}{0.23\textwidth}
		\includegraphics[width=\linewidth]{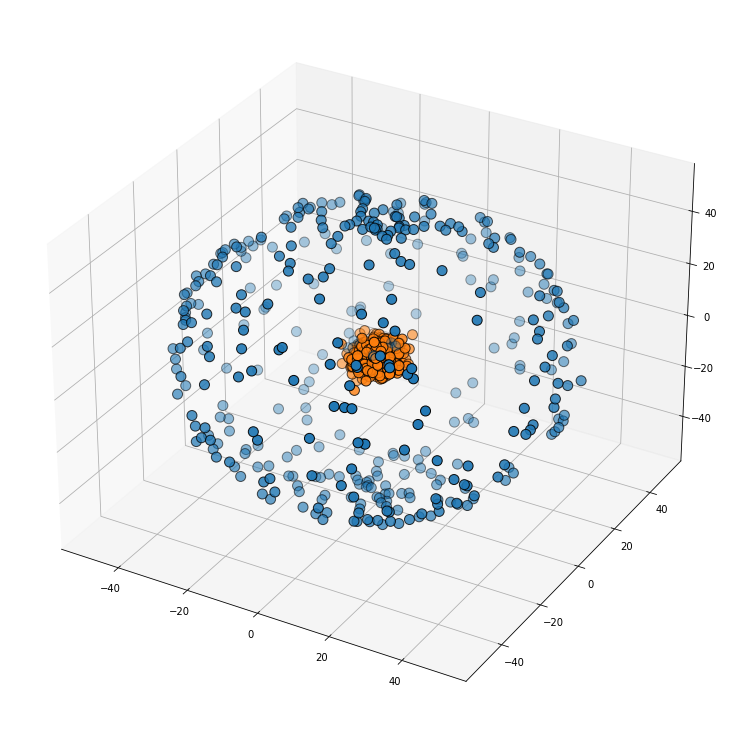}
		\caption{Atom}
	\end{subfigure}
	\hfill
	\begin{subfigure}{0.23\textwidth}
		\includegraphics[width=\linewidth]{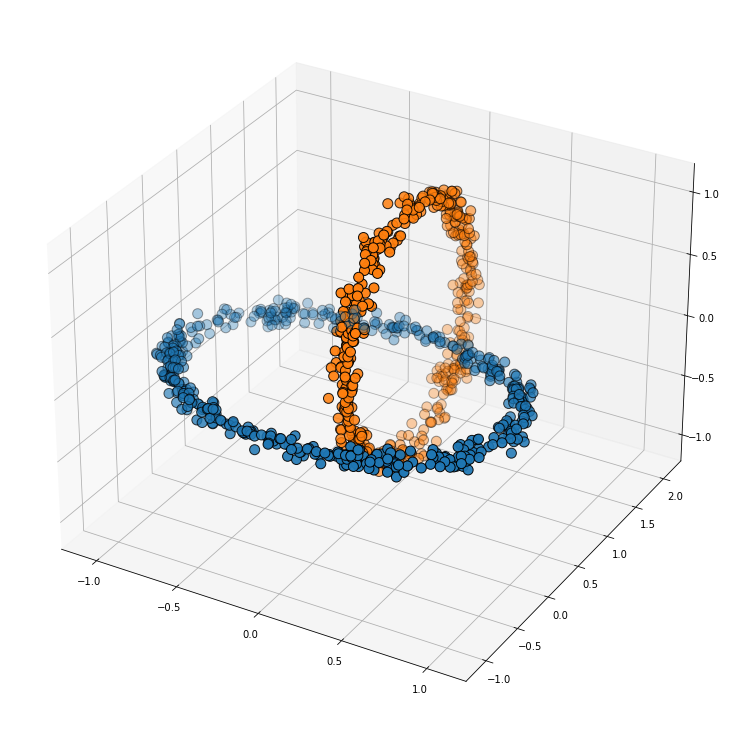}
		\caption{Chainlink}
	\end{subfigure}
	\hfill
	\begin{subfigure}{0.23\textwidth}
		\includegraphics[width=\linewidth]{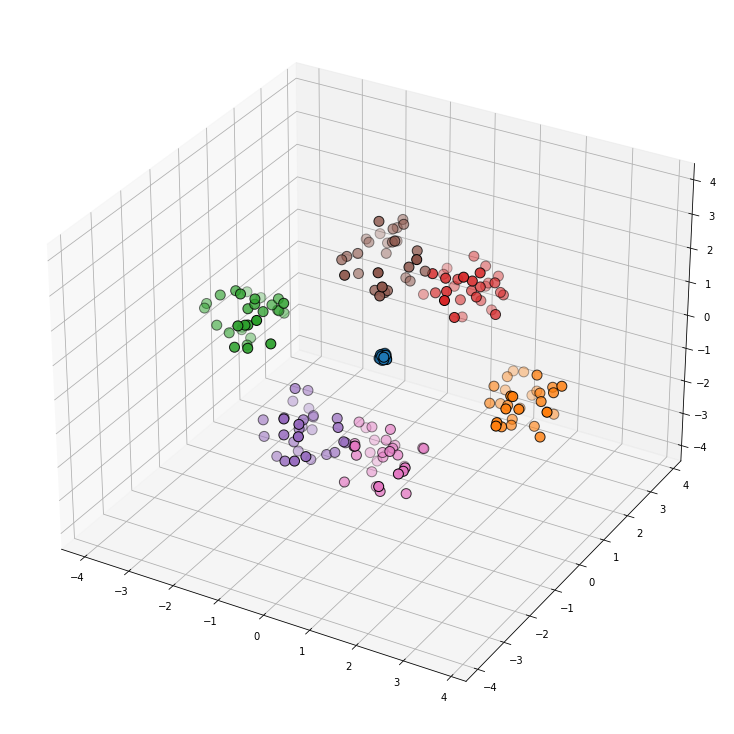}
		\caption{Hepta}
	\end{subfigure}
	\hfill
	\begin{subfigure}{0.23\textwidth}
		\includegraphics[width=\linewidth]{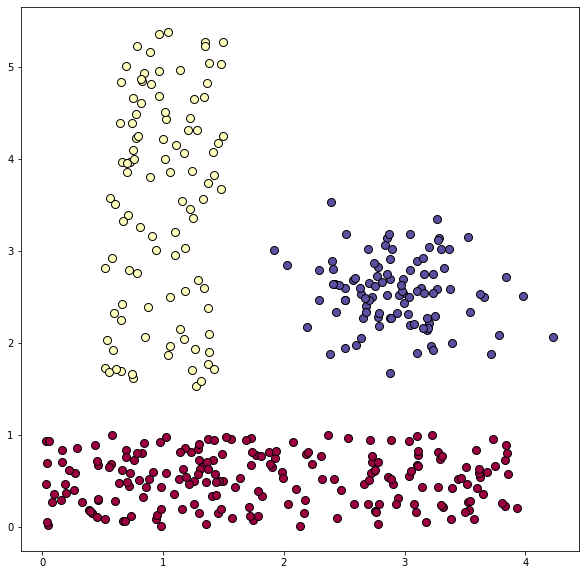}
		\caption{Lsun}
	\end{subfigure}
	
	\vspace{0.5em}
	
	\begin{subfigure}{0.23\textwidth}
		\includegraphics[width=\linewidth]{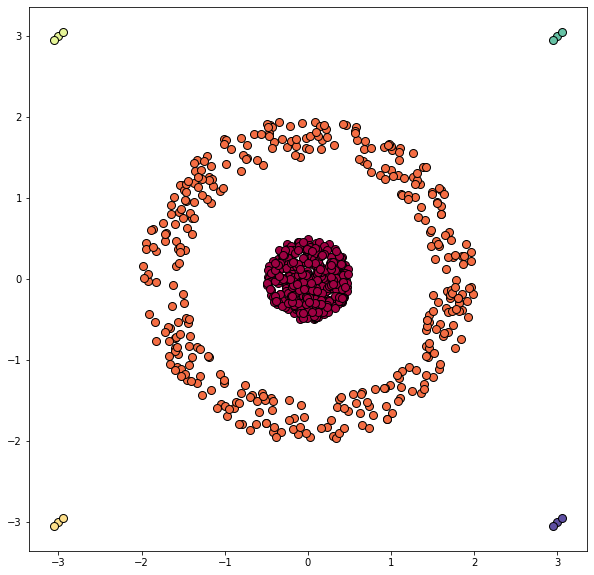}
		\caption{Target}
	\end{subfigure}
	\hfill
	\begin{subfigure}{0.23\textwidth}
		\includegraphics[width=\linewidth]{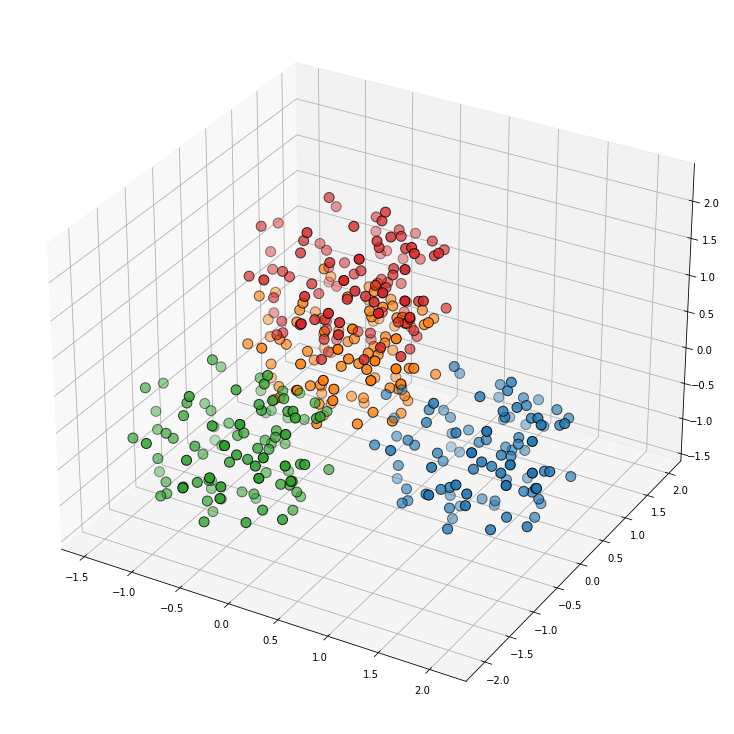}
		\caption{Tetra}
	\end{subfigure}
	\hfill
	\begin{subfigure}{0.23\textwidth}
		\includegraphics[width=\linewidth]{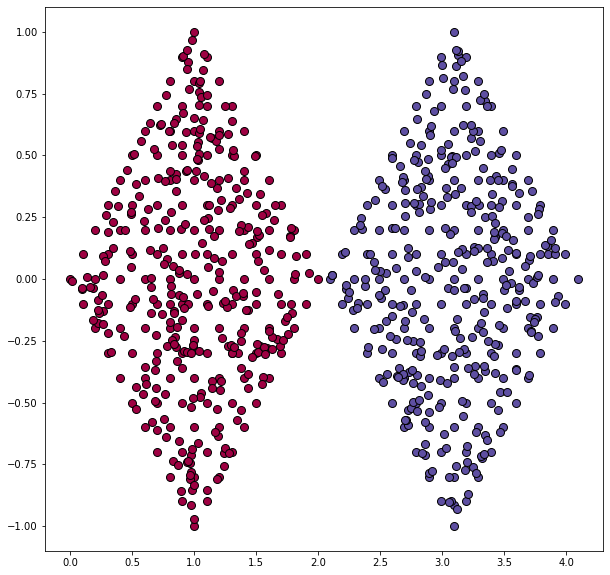}
		\caption{TwoDiamonds}
	\end{subfigure}
	\hfill
	\begin{subfigure}{0.23\textwidth}
		\includegraphics[width=\linewidth]{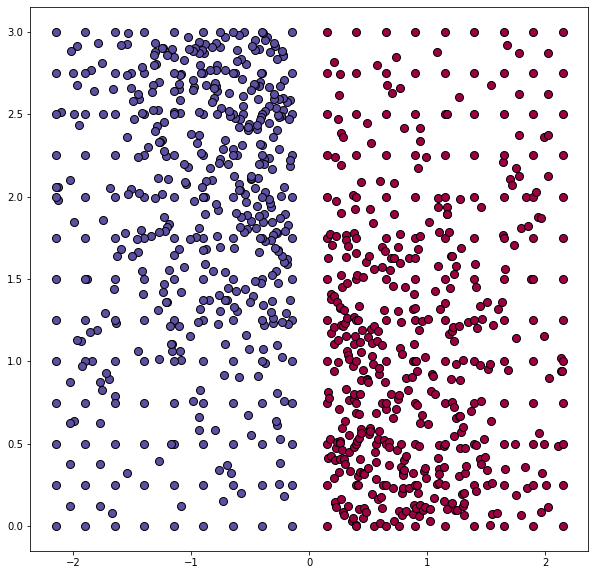}
		\caption{Wingnut}
	\end{subfigure}
	
	\caption{
		Overview of the FCPS datasets used in this paper, shown with ground-truth labels. These diverse cluster shapes—including rings, overlapping spheres, and complex symmetric forms—serve as a robust benchmark for evaluating radius-guided post-processing.
	}
	\label{fig:fcps-datasets}
\end{figure}

Figure~\ref{fig:fcps-datasets} shows the FCPS datasets used in our experiments, each displayed with its ground-truth labels. The diversity of cluster structures—ranging from thin interlocking shapes to dense symmetric blobs—offers a robust testbed for the algorithm's flexibility and generality.

\subsection{FCPS Benchmark Results of Post-Processing Compared to k-Means}

The FCPS suite includes a variety of 2D and 3D clustering challenges with ground-truth labels. One dataset, FCPS Engytime, was excluded from evaluation, as it contains overlapping clusters for which this algorithm is not a natural fit. We first ran standard k-means with $k$ equal to the ground-truth number of clusters, and then ran each experiment again with an intentionally overestimated $k$, applying the radius-based post-processing step.

Each experiment was repeated 100 times, and the best of five runs was kept. Tables~\ref{tab:fcps-pre} and~\ref{tab:fcps-post} summarize the results, reporting median clustering success and interquartile range across runs.

\begin{table}[htbp]
	\centering
	\begin{tabular}{|l|c|c|c|}
		\hline
		\textbf{Dataset} & \textbf{$k$ (Ground Truth)} & \textbf{Pre-Merge Median Success} & \textbf{IQR (Pre)} \\
		\hline
		FCPS Atom        & 2   & 0.7038   & 0.0287    \\
		FCPS Chainlink   & 2   & 0.653   & 0.006     \\
		FCPS Hepta       & 7   & \cellcolor{gray!20}1.0   & 0.0     \\
		FCPS Lsun        & 3   & 0.765   & 0.0125     \\
		FCPS Target      & 6   & 0.6338   & 0.0091    \\
		FCPS Tetra       & 4   & \cellcolor{gray!20}1.0   & 0.0     \\
		FCPS TwoDiamonds & 2   & \cellcolor{gray!20}1.0   & 0.0     \\
		FCPS Wingnut     & 2   & \cellcolor{gray!20}0.9636   & 0.0     \\
		\hline
	\end{tabular}
	\caption{Baseline: Clustering performance of standard k-means on the FCPS dataset suite. Median success and interquartile range are based on multiple runs. Cells with a light gray background indicate median success values above 0.8.}
	\label{tab:fcps-pre}
\end{table}

\begin{table}[htbp]
	\centering
	\begin{tabular}{|l|c|c|c|}
		\hline
		\textbf{Dataset} & \textbf{Post-Proc $k$ Range, Value} & \textbf{Post-Merge Median Success} & \textbf{IQR (Post)} \\
		\hline
		FCPS Atom        & $15 \leq k \leq 20,\ k=20$  & \cellcolor{gray!20}1.0 & 0.0 \\
		FCPS Chainlink   & $15 \leq k \leq 20,\ k=16$  & \cellcolor{gray!20}1.0 & 0.177 \\
		FCPS Hepta       & $10 \leq k \leq 20,\ k=20$  & \cellcolor{gray!20}1.0 & 0.0 \\
		FCPS Lsun        & $15 \leq k \leq 20,\ k=20$  & \cellcolor{gray!20}1.0 & 0.075 \\
		FCPS Target      & $15 \leq k \leq 40,\ k=17$  & \cellcolor{gray!20}0.9143 & 0.2172 \\
		FCPS Tetra       & $k = 4$              & \cellcolor{gray!20}1.0 & 0.0 \\
		FCPS TwoDiamonds & $10 \leq k \leq 20,\ k=10$  & \cellcolor{gray!20}1.0 & 0.0 \\
		FCPS Wingnut     & $65 \leq k \leq 75,\ k=70$  & \cellcolor{gray!20}0.9823 & 0.4870 \\
		\hline
	\end{tabular}
	\caption{Clustering performance after radius-based post-processing. The ranges for $k$ give similar results. Cells with a light gray background indicate median success values above 0.8.}
	\label{tab:fcps-post}
\end{table}

Tetra was the only dataset for which any $k \neq 4$ failed to recover the correct clusters. The dataset consists of four nearly touching blobs in 3D space—an ideal case for standard k-means with $k = 4$, as the clusters are nearly spherical. With this value, post-processing neither helped nor hurt. But for higher $k$, the fact that \textit{the inter-cluster distance between neighboring points is close to the intra-cluster distance} causes most radii to overlap, resulting in the trivial cluster after merging.

The results for the datasets except Tetra show that radius-based post-processing consistently outperforms pure k-means (4 out of 7 datasets), or at the very least relaxes the need to know $k$ precisely in advance where k-means already achieved good clustering results, given the correct $k$ (3 out of 7 datasets).

\subsection{FCPS Benchmark Results for Partitioning the Data Space}

To test the method’s ability to support distributed or large-scale clustering, we evaluated to which degree it recovers global structure after dividing the feature space into equal rectangular tiles. This simulates a setting where clustering is performed locally within each tile, followed by a global merge phase using the radius-based overlap rule described earlier.

We applied this test to all previously used FCPS datasets. Each dataset was split into four quadrants by partitioning the feature space in half along the first two dimensions. Local k-means clustering and post-processing were performed independently within each tile, with $k$ set as a percentage of the tile’s point count. A final global merge step was then applied across tile boundaries.

The method showed strong recovery in most datasets (Atom, Hepta, Lsun, TwoDiamonds, Wingnut, Target, 6 out of 8), as shown in table~\ref{tab:fcps-partitioned}. This indicates that the method can succeed in many partitioned settings, even without global awareness during the initial clustering phase.

In Chainlink, the interlocking rings are thin and sparsely sampled. After partitioning, the fragments were too small to reconnect, and the method typically returned 5–12 clusters despite the ground truth being 2.

For Tetra the algorithm in this stage can't rely on k-means doing all the work, because the fragile figure (on which k-means performed well) has been cut into four independent pieces, and that shows in the result. 

Local $k$ values based on percentages, like global $k$, are forgiving when combined with post-processing. But there is no universal default. Domain knowledge or evaluation on labeled subsets is often needed to guide local choices. In unlabeled settings, one alternative is to run the method multiple times with varying $k$ values or random seeds and look for stability across runs (see Section~\ref{sec:post-processing}). 	

\begin{table}[htbp]
	\centering
	\begin{tabular}{|l|c|c|c|c|}
		\hline
		\textbf{Dataset} & \textbf{$k$ (\% of tile points)} & \textbf{Median Success (Reass.)} & \textbf{IQR (Reassembled)} \\
		\hline
		FCPS Atom        & 5.0   & \cellcolor{gray!20}1.0  & 0.0138 \\
		FCPS Chainlink   & 5.0   & 0.297  & 0.0963 \\
		FCPS Hepta       & 15.0   & \cellcolor{gray!20}0.9811  & 0.0142 \\
		FCPS Lsun        & 5.0   & \cellcolor{gray!20}0.995  & 0.075 \\
		FCPS Target      & 3.0   & \cellcolor{gray!20}0.8429  & 0.0065 \\
		FCPS Tetra       & 20.0  & 0.7038  & 0.2325 \\
		FCPS TwoDiamonds & 5.0   & \cellcolor{gray!20}1.0  & 0.0 \\
		FCPS Wingnut     & 8.0   & \cellcolor{gray!20}0.9951  & 0.0367 \\
		\hline
	\end{tabular}
	\caption{Clustering performance on FCPS datasets after partitioning the feature space into four tiles. Local $k$ is the percentage of data points per tile used for initial clustering. Median success and interquartile range reflect performance after global radius-based merging. Cells with a light gray background indicate median success values above 0.8.}
	\label{tab:fcps-partitioned}
\end{table}

The high recovery rate in most datasets supports the use of this method in tiled or distributed environments, where data must be processed locally due to memory constraints, streaming inputs, or parallelization demands. It also opens the door to recursive refinement strategies, making the method a viable candidate for scalable hierarchical clustering pipelines.

Table~\ref{tab:fcps-summary} summarizes the core findings of both experiments, highlighting the success of radius-based post-processing and its limitations in certain scenarios.

\begin{table}[htbp]
	\centering
	\begin{tabular}{|l|c|c|c|c|}
		\hline
		\textbf{Dataset} & \textbf{Baseline} & \textbf{Radius-Based Merge} & \textbf{Radius-Based Merge (Split Dataset)} \\
		\hline
		FCPS Atom        & 0.7038   & 1.0   & 1.0 \\
		FCPS Chainlink   & 0.653   & 1.0   & \cellcolor{yellow!20}0.297 \\
		FCPS Hepta       & 1.0   & 1.0   & 0.9811 \\
		FCPS Lsun        & 0.765   & 1.0   & 0.995 \\
		FCPS Target      & 0.6338    & 0.9143  & 0.8429 \\
		FCPS Tetra       & 1.0 & \cellcolor{yellow!20}1.0  & \cellcolor{yellow!20}0.7038 \\
		FCPS TwoDiamonds & 1.0   &  1.0 & 1.0 \\
		FCPS Wingnut     & 0.9636  & 0.9823 & 0.9951 \\
		\hline
	\end{tabular}
	\caption{Combined summary of the experimental results. Values represent the median clustering success in recovering ground-truth labels: baseline k-means, radius-based post-processing, and radius-based post-processing on partitioned data, assembling the local results to a combined clustering. Results where post-processing or reassembly faced problems are marked in yellow and discussed in the text.}
	\label{tab:fcps-summary}
\end{table}

\section{Discussion and Conclusion}

This paper introduced a radius-based post-processing enhancement to the k-means clustering algorithm. By assigning each cluster a radius—defined as the distance to its farthest point—and merging clusters whose radii overlap, the method addresses two common limitations of k-means: sensitivity to the choice of $k$, and suboptimal performance on non-convex shapes.

The approach requires no additional tuning parameters beyond the initial $k$, which can be set conservatively high. The merge rule is simple, geometric, and interpretable. Because it operates entirely as a post-processing step, the method remains compatible with existing implementations.

Empirically, the method achieves high clustering accuracy on the FCPS dataset suite. It consistently recovers ground-truth clusters with high accuracy.

The method's design also supports scalability through recursive partitioning. By dividing the feature space into smaller tiles and clustering them independently, it enables distributed or parallel processing while preserving global structure during the merge phase. This makes the approach suitable for large-scale, spatially distributed, or streaming datasets.

\subsection{Limitations and Future Work}

Despite its strengths, the method has several limitations. It assumes that cluster \textit{extents}—not necessarily the clusters themselves—can be approximated by circles that connect to other extents of the same cluster without connecting to extents of other clusters. In datasets with elongated, curved, or irregularly dense clusters that are close to each other, this simplification may cause the merge rule to fail in reconnecting meaningful structure. The method also depends on an over-partitioning step via standard k-means, which may still suffer from poor initialization.

Future work may explore alternative radius definitions (e.g., average or quantile-based distance, or scaling), adaptive merging heuristics, or confidence scores for proposed merges. Extending the method to high-dimensional data and formally characterizing its behavior under noise are also open directions.

Although the method is not designed for high-noise environments, preliminary results suggest that small, isolated clusters (e.g., those with radius zero, or a single member) may be filtered out with simple heuristics. A more principled noise-handling extension remains an avenue for future development.

\subsection{Conclusion}

The radius-based merging method offers a conceptually simple yet effective enhancement to k-means. It improves shape sensitivity, enables post hoc correction of over-segmentation, and thereby allows intentional over-segmentation to relax the need for an exact choice of $k$. It also achieves high clustering accuracy on many partitioned datasets, which makes it especially relevant for big data or parallelized clustering pipelines. Its geometric intuition, ease of implementation, and good empirical performance suggest that it can serve as a useful drop-in refinement for clustering pipelines that rely on k-means as a first step.

\bibliographystyle{plain}  
\bibliography{radius_guided_kmeans.bbl}

\end{document}